\documentclass[11pt,a4paper]{article}
\usepackage{acl2015}
\usepackage{times}
\usepackage{latexsym}
\usepackage{url}
\usepackage{color}
\usepackage{multirow}
\usepackage{comment}
\usepackage{amsmath} 
\usepackage{amssymb}

\usepackage[font=normalsize,skip=2pt]{caption}

\usepackage{etoolbox}
\usepackage{graphicx}
 
 \usepackage{enumitem}
  {\begin{itemize}[topsep=0pt, partopsep=0pt] \footnotesize%
    \setlength{\itemsep}{0pt}%
    \setlength{\parskip}{0pt}%
    }%
  {\end{itemize}}
  
  \usepackage{color}
  {\begin{enumerate}≈ \footnotesize%
    \setlength{\itemsep}{0pt}%
    \setlength{\parskip}{0pt}%
    }%
  {\end{enumerate}}
  
  \usepackage{CJKutf8}

\author{Jiwei Li$^1$, Minh-Thang Luong$^1$, Dan Jurafsky$^1$ and Eduard Hovy$^2$
\\$^1$Computer Science Department, Stanford University, Stanford, CA 94305\\
$^2$Language Technology Institute, Carnegie Mellon University, Pittsburgh, PA 15213\\jiweil,lmthang,jurafsky@stanford.edu ~~~~~~~ehovy@andrew.cmu.edu
}

\title{When Are Tree Structures Necessary for Deep Learning of Representations?}

\begin{document}
\maketitle
\begin{abstract}
Recursive neural models, which use syntactic parse trees
to recursively generate representations bottom-up, are a popular architecture.
But there have not been rigorous evaluations showing for
exactly which tasks this syntax-based method is appropriate.
In this paper we benchmark {\bf recursive} neural models
against sequential {\bf recurrent} neural models (simple recurrent and LSTM models),
enforcing apples-to-apples comparison as much as possible. 
We investigate 4 tasks:
(1) sentiment classification at the sentence level and phrase level;
(2) matching questions to answer-phrases;
(3) discourse parsing;
(4) semantic relation extraction (e.g., {\em component-whole} between nouns).

Our goal
is to understand better when, and why, recursive models can outperform 
simpler models.  
We find that recursive models help mainly
on tasks (like semantic relation extraction) that require associating headwords 
across a long distance, particularly on very long sequences.
We then introduce a method for allowing recurrent models to achieve 
similar performance: breaking long sentences into  clause-like units
at punctuation and processing them separately before combining.
Our results thus help understand the limitations of both classes of models,
and suggest directions for improving recurrent models.
\end{abstract}

\setlength{\textfloatsep}{5pt}

\section{Introduction}
Deep learning based methods learn low-dimensional, real-valued vectors for word
tokens, mostly from large-scale data corpus (e.g.,
\cite{mikolov2013linguistic,le2014distributed,collobert2011natural}),
successfully capturing syntactic and semantic aspects of text.

For tasks where the inputs are larger text units 
(e.g., phrases, sentences or documents), a compositional model
is first needed to aggregate  tokens into a  vector with fixed
dimensionality  that can be used as a feature for other NLP tasks.
Models for achieving this usually fall into two categories:
{\bf recurrent} models and {\bf recursive} models:

{\bf Recurrent} models (also referred to as {\bf sequence} models) deal successfully with time-series data \cite{pearlmutter1989learning,dorffner1996neural} like speech 
\cite{robinson1996use,lippmann1989review,graves2013speech} or handwriting recognition \cite{graves2009offline,graves2012supervised}.
They were applied early on to NLP \cite{elman1990finding},
modeling a sentence as tokens processed sequentially,
at each step combining the current token with previously built embeddings.
Recurrent models can be extended to bidirectional ones from both left-to-right and right-to-left.  
These models generally consider no linguistic structure aside from word order. 

{\bf Recursive} neural models (also referred to as {\bf tree} models), by contrast,
are structured by syntactic parse trees.
Instead of considering tokens sequentially, recursive models combine neighbors based on the 
recursive structure of parse trees,
starting from the leaves and proceeding recursively in a bottom-up fashion until the root of the parse tree is reached.
For example, for the phrase {\it the food is delicious}, following the operation sequence
{\it ( (the food) (is delicious) )} rather than 
the sequential order {\it (((the food) is) delicious)}. 
Many recursive models  have been proposed (e.g., \cite{paulus2014global,irsoy2014deep}),
and applied to various NLP tasks, among them
entailment \cite{bowman2013can,bowman2014recursive},
sentiment analysis \cite{socher2013recursive,irsoy2013bidirectional,dong2014adaptive},
question-answering \cite{iyyer2014neural}, relation classification \cite{socher2012semantic,hashimoto2013simple}, 
and discourse \cite{limodel}.

One possible advantage of recursive models is their 
%
potential for capturing
long-distance dependencies: two tokens may be structurally close to each other, even though they are far away in word sequence.
For example, a verb and its corresponding direct object can be far away in terms of tokens if many adjectives lies in between, but they are adjacent in the parse tree \cite{irsoy2013bidirectional}.
But we don't know 
if this advantage is truly important, and if so for which tasks,
or whether other issues are at play.
Indeed, the reliance of recursive models on parsing is also a potential disadvantage, given
that parsing is relatively slow, domain-dependent, and can be errorful.


On the other hand, recent progress in multiple subfields of neural NLP has suggested that recurrent nets
may be sufficient to deal with many of the tasks for which recursive models have been proposed.
Recurrent models without parse structures have shown good results in
sequence-to-sequence generation \cite{sutskever2014sequence} for machine translation (e.g., \cite{kalchbrenner2013recurrent,bahdanau2014neural,luong2014addressing}), parsing \cite{vinyals2014grammar}, and
sentiment, where for example recurrent-based paragraph vectors \cite{le2014distributed} 
outperform recursive models \cite{socher2013recursive} on the Stanford sentiment-bank dataset.

Our goal in this paper is thus to investigate a number of tasks
with the goal of understanding for which kinds of problems recurrent models may be sufficient,
and for which  kinds recursive models offer specific advantages.
We investigate four tasks with different properties.  
\begin{itemize}
\item Binary {\bf sentiment classification}  at the sentence level
\cite{pang2002thumbs} and phrase level \cite{socher2013recursive} 
that focus on understanding the role of recursive models in dealing with semantic compositionally in various scenarios such as different lengths of inputs and whether or not supervision is comprehensive.
\item 
{\bf Phrase Matching} on the UMD-QA dataset \cite{iyyer2014neural} can help see 
the difference between
outputs from 
 intermediate components from different models, i.e., representations for intermediate parse tree nodes and outputs from recurrent models at different time steps. 
It also helps see
whether
parsing is useful for finding similarities between question sentences and target phrases.
\item   
 {\bf Semantic Relation Classification}
on the SemEval-2010 \cite{hendrickx2009semeval} data can help understand whether parsing is helpful in
dealing with long-term dependencies, such as relations between two words that are far apart in the sequence. 
\item 
{\bf Discourse parsing}
(RST dataset) is useful for measuring the extent to which parsing improves discourse tasks
that need to combine meanings of larger text units.  Discourse parsing treats
elementary discourse units (EDUs) as basic units to operate on, which are usually short clauses. The task also sheds light on the extent to which syntactic structures help acquire shot text representations. 
\end{itemize}

The  principal motivation for this paper
is to understand better when, and why, recursive models are needed to outperform simpler models by enforcing 
apples-to-apples comparison as much as possible.
This paper applies existing models to existing tasks,
barely offering novel algorithms or tasks.
Our goal is rather an analytic one, to investigate different versions of recursive and recurrent models.   
This work helps understand the limitations of both classes of models, and suggest directions for improving recurrent models.

The rest of this paper organized as follows: We detail versions of recursive/recurrent models in Section 2, present the tasks and results in Section 3, and
conclude with discussions in Section 4.

\section{Recursive and Recurrent Models}
\subsection{Notations}
We assume that the text unit $S$, which could be a phrase, a sentence
or a document, is comprised of a sequence of tokens/words: $S=\{w_1,w_2, ..., w_{N_S}\}$, 
where $N_s$ denotes the number of tokens in $S$.
Each word w is associated with a K-dimensional vector embedding
$e_w=\{e_w^1,e_w^2,...,e_w^K\}$. The goal of  recursive and recurrent
models is to map the sequence to a K-dimensional $e_S$, based on
its tokens and their correspondent embeddings.

\paragraph{Standard Recurrent/Sequence Models} 
A recurrent network successively takes word $w_t$ at step $t$,
combines its vector representation $e_{t}$ with the previously built
hidden vector $h_{t-1}$  from time $t-1$, calculates the resulting
current embedding $h_t$, and passes it to the next step.  The embedding
$h_t$ for the current time $t$ is thus:
\begin{equation}
h_{t}=f(W\cdot h_{t-1}+V\cdot e_{t})
\end{equation}
where $W$ and $V$ denote compositional matrices. 
If $N_s$ denotes the length of the sequence, $h_{N_s}$ represents
the whole sequence $S$. 
\paragraph{Standard recursive/Tree models}
Standard recursive models work in a similar way,
but processing neighboring words
by parse tree order rather than sequence order.
It computes a
representation for each parent node based on its immediate
children recursively in a bottom-up fashion until reaching the root
of the tree. For a given node $\eta$ in the tree and
its left child
$\eta_{\text{left}}$ (with  representation $e_{{\text{left}}}$) and right 
child $\eta_{\text{right}}$ (with  representation $e_{{\text{right}}}$),
the standard recursive network calculates $e_{\eta}$ as follows:
\begin{equation}
e_{\eta}=f(W\cdot e_{\eta_{\text{left}}}+V\cdot e_{\eta_{\text{right}}})
\end{equation}

\paragraph{Bidirectional Models} \cite{schuster1997bidirectional}  add 
bidirectionality to the recurrent framework where embeddings for each time are calculated both forwardly and backwardly:
\begin{equation}
\begin{aligned}
&h_{t}^{\rightarrow}=f(W^{\rightarrow}\cdot h_{t-1}^{\rightarrow}+V^{\rightarrow}\cdot e_{t}) \\
&h_{t}^{\leftarrow}=f(W^{\leftarrow}\cdot h_{t+1}^{\leftarrow}+V^{\leftarrow}\cdot e_{t})
\end{aligned}
\end{equation}
Normally, final representations for sentences can be achieved either by concatenating vectors calculated from both directions
$[e_1^{\leftarrow},e_{N_S}^{\rightarrow}]$  or using further compositional operation  to preserve  vector dimensionality
\begin{equation}
\begin{aligned}
&h_t=f(W_L\cdot [h_t^{\leftarrow},h_t^{\rightarrow}])\\
\end{aligned}
\end{equation}
where $W_L$ denotes a $K\times 2K$ dimensional matrix.

\paragraph{Long Short Term Memory (LSTM)}
LSTM models \cite{hochreiter1997long} are defined as follows: 
given a sequence of inputs  $X=\{x_1,x_2,...,x_{n_X}\}$, an LSTM associates each timestep with an input, memory and output gate, 
respectively denoted as $i_t$, $f_t$ and $o_t$.
We notationally disambiguate $e$ and $h$: $e_t$ denotes the vector for individual text units (e.g., word or sentence) at time step t, while $h_t$ denotes the vector computed by the LSTM model at time t by combining $e_t$ and $h_{t-1}$. 
$\sigma$ denotes the sigmoid function. The vector representation $h_t$ for each time-step $t$ is given by:

\begin{equation}
\Bigg[
\begin{array}{lr}
i_t\\
f_t\\
o_t\\
l_t\\
\end{array}
\Bigg]=
\Bigg[
\begin{array}{c}
\sigma\\
\sigma\\
\sigma\\
\text{tanh}\\
\end{array}
\Bigg]
W\cdot
\Bigg[
\begin{array}{c}
h_{t-1}\\
e_{t}\\
\end{array}
\Bigg]
\end{equation}
\begin{equation}
c_t=f_t\cdot c_{t-1}+i_t\cdot l_t\\
\end{equation}
\begin{equation}
h_{t}^s=o_t\cdot c_t
\end{equation}
where $W\in \mathbb{R}^{4K\times 2K}$. Labels at the phrase/sentence level are predicted representations outputted from the last time step. 

\paragraph{Tree LSTMs}
Recent research has extended the LSTM idea to tree-based structures \cite{zhu2015long,tai2015improved} that associate memory and forget gates to nodes of the parse trees. 

\paragraph{Bi-directional LSTMs} These combine bi-directional models and LSTMs. 

\section{Experiments}
In this section, we detail our experimental settings and results.
We consider the following tasks, each representative of a different class of NLP tasks.

\noindent {\bf 1.} {\bf Binary sentiment classification} on the \newcite{pang2002thumbs} dataset.
This addresses the issues where supervision only appears globally after a long sequence of operations.

\noindent {\bf 2.} {\bf Sentiment Classification on the Stanford Sentiment Treebank} \cite{socher2013recursive}: comprehensive labels are found for words and phrases where local compositionally (such as from negation, mood, or others cued by phrase-structure) is to be learned.

\noindent {\bf 3.} {\bf Sentence-Target Matching} on the  UMD-QA dataset \cite{iyyer2014neural}: Learns matches between target and 
components in the source sentences, which are  parse tree nodes for recursive models and different time-steps for recurrent models.

\noindent {\bf 4.} {\bf Semantic Relation Classification} on the SemEval-2010 task \cite{hendrickx2009semeval}.
Learns long-distance relationships  between two words that may be far apart sequentially.

\noindent {\bf 5.} {\bf Discourse Parsing} \cite{li2014recursive,hernault2010hilda}: Learns sentence-to-sentence relations based 
on calculated representations.

In each case we followed the protocols described in the original papers. 
We first group the algorithm variants into two groups as follows:
\begin{itemize}
\item Standard tree models vs standard sequence models vs standard bi-directional sequence models
\item LSTM tree models, LSTM sequence models vs LSTM bi-directional sequence models. 
\end{itemize}
We employed standard training frameworks for neural models: for
each task, we used stochastic gradient decent using AdaGrad
\cite{duchi2011adaptive} with minibatches \cite{cotter2011better}.
Parameters are tuned using the development dataset if available in
the original datasets or from cross-validation if not.  Derivatives
are calculated from standard back-propagation \cite{goller1996learning}.
Parameters to tune include size of mini batches, learning rate,
and parameters for L2 penalizations.  The number of running iterations is
treated as a parameter to tune and the model achieving best performance
on the development set is used as the final model to be evaluated.

For settings where no repeated experiments are performed, 
the bootstrap test
is adopted for statistical significance testing \cite{efron1994introduction}. 
Test scores that achieve significance level of 0.05 are marked by an asterisk (*).

\subsection{Stanford Sentiment TreeBank}
\begin{figure*}
\centering
\includegraphics[width=4.5in]{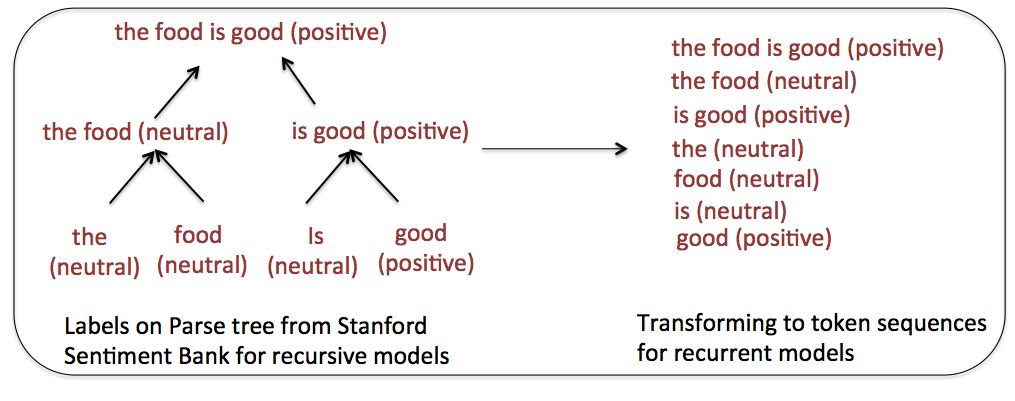}
\caption{Transforming Stanford Sentiment Treebank to Sequences for Sequence Models. }
\label{transform}
\end{figure*}
\paragraph{Task Description}
We start with the Stanford Sentiment TreeBank \cite{socher2013recursive}.
This dataset contains gold-standard labels for every parse tree constituent, from the sentence to phrases to individual words.

Of course, any conclusions drawn from 
implementing sequence models on a  dataset  that was based on
parse trees may have to be weakened,
since sequence models may still benefit from the way that 
the dataset was collected.  Nevertheless we add an evaluation on
this dataset because it has been a widely used benchmark dataset for neural model evaluations. 

For recursive models, we followed the protocols in \newcite{socher2013recursive} where
node embeddings in the parse trees are obtained from recursive models and then fed to a softmax classifier.
We transformed the
dataset for recurrent model use as illustrated in Figure~\ref{transform}.
Each phrase is reconstructed from parse tree nodes and treated  as a separate data point.  
As
the treebank contains 11,855 sentences with 215,154 phrases, the
reconstructed dataset for recurrent models comprises 215,154
examples.
Models are evaluated at both the phrase level (82,600 instances) and the sentence root level (2,210 instances).

\begin{table}[!ht]
\center
\small
\begin{tabular}{l|l|l}
&Fine-Grained&Binary\\\hline
Tree&0.433&0.815\\\hline\hline
Sequence&0.420 (-0.013)&0.807 (-0.007) \\
P-value&0.042*&0.098\\\hline\hline
Bi-Sequence&0.435 (+0.08) &0.816 (+0.002)\\
P-value&0.078&0.210\\\hline\hline
\end{tabular}
\caption{Test set accuracies on the Stanford Sentiment
Treebank at root level.}
\end{table}

\begin{table}[!ht]
\center
\small
\begin{tabular}{l|l|l}
&Fine-Grained&Binary\\\hline
Tree&0.820&0.860\\\hline\hline
Sequence&0.818 (-0.002)&0.864 (+0.004) \\
P-value&0.486&0.305\\\hline\hline
Bi-Sequence&0.826 (+0.06) &0.862 (+0.002)\\
P-value&0.148&0.450\\\hline\hline
\end{tabular}
\caption{Test set accuracies on the Stanford Sentiment
Treebank  at phrase level.}
\end{table}
Results are shown in Table 1 and 2\footnote{The performance of our
implementations of recursive models is not exactly identical to
that reported in \newcite{socher2013recursive}, but the relative
difference is around $1\%$ to $2\%$. }.
When comparing the standard version of tree models to sequence models, 
we find it helps a bit at root level identification (for sequences
but not bi-sequences), but yields no significant improvement at the phrase level. 

\begin{figure}
\centering
\includegraphics[width=2in]{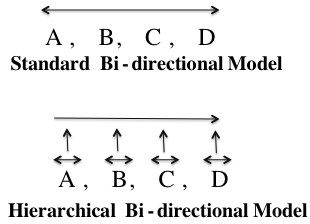}
\caption{Illustration of two sequence models. A, B, C, D denote clauses or sub sentences separated by punctuation.}
\label{Illustration}
\end{figure}

\paragraph{LSTM}
\newcite{tai2015improved} discovered that LSTM tree models generate better performances in terms of sentence {\bf root} level evaluation
  than sequence models. We explore this task a bit more by training deeper and more sophisticated models. We examine the following three models:
\begin{enumerate}
\item Tree-structured LSTM models \cite{tai2015improved}\footnote{Tai et al.. achieved 0.510 accuracy in terms of fine-grained evaluation at the root level as reported in \cite{tai2015improved}, similar to results from our implementations (0.504). }.
\item Deep Bi-LSTM sequence models (denoted as {\bf Sequence}) that treat the whole sentence as just one sequence.
\item Deep Bi-LSTM hierarchical sequence models (denoted as {\bf Hierarchical Sequence}) that first slice the sentence into a sequence of sub-sentences by using a look-up table of punctuations (i.e., comma, period, question mark and exclamation mark).
The representation for each sub-sentence is first computed  separately,
and another level of sequence LSTM (one-directional) is then used to join the sub-sentences.
Illustrations are shown in Figure\ref{Illustration}. 
\end{enumerate}

We consider the third model because the dataset
used in \newcite{tai2015improved} contains long sentences and the
evaluation is performed only at the sentence root level.  Since a parsing
algorithm will naturally break long sentences into sub-sentences,
we'd like to know whether any performance boost is introduced by the
intra-clause parse tree structure or just by
this broader segmentation of a sentence into clause-like units;
this latter advantage could be approximated
by using punctuation-based approximations to clause boundaries.

We run 15 iterations for each algorithm. Parameters are harvested at the end of each iteration; those performing best on the dev set
are used on the test set. The whole process takes roughly 15-20 minutes on a single GPU machine\footnote{Tesla K40m, 2880 Cuda cores.}.
For a more convincing comparison, we did not use the bootstrap test where parallel examples are generated from one same dataset. Instead, 
we repeated the aforementioned procedure for each algorithm 20 times 
and report accuracies  with standard deviation in Table \ref{table}. 
\begin{table}[!ht]
\centering
\small
\begin{tabular}{c|c|c|c}
Model&all-fine&root-fine&root-coarse\\\hline
Tree LSTM &{\bf 83.4} (0.3) &50.4 (0.9) &86.7 (0.5) \\
Bi-Sequence&83.3 (0.4) &49.8 (0.9)&86.7 (0.5) \\
Hier-Sequence&82.9 (0.3)&{\bf 50.7} (0.8) &{\bf 86.9} (0.6)  \\\hline
\end{tabular}
\caption{Test set accuracies on the Stanford Sentiment
Treebank with deviations.  For our experiments, we report
accuracies over 20 runs with standard deviation.}
\label{table}
\end{table}

Tree LSTMs are equivalent or marginally better than
standard bi-directional sequence model (two-tailed p-value equals 0.041*,
and only at the root level, with p-value for the phrase level at 0.376).  The hierarchical sequence model
achieves the same performance with a p-value of 0.198. 
\paragraph{Discussion}
The results above suggest that clausal segmentation
of long sentences offers a slight performance boost,
a result also supported by the fact that 
very little difference exists between the three models for 
phrase-level sentiment evaluation. 
Clausal segmentation
of long sentences thus provides a simple approximation 
to parse-tree based models. 

We suggest a few reasons for
this slightly better performances introduced by clausal segmentation:
\begin{enumerate}
\item Treating clauses as basic units (to the extent that punctuation
approximates clauses) preserves the semantic structure of text.
\item
Semantic compositions such as negations or conjunctions usually appear at the clause level.
Working on clauses individually  and then combining them model
inter-clause compositions.
\item 
Errors are back-propagated to individual tokens using
fewer steps in hierarchical models than in standard models. 
Consider a movie review ``simple as the plot was , i still like it a lot".
With standard  recurrent models
it takes 12 steps before the prediction error gets back to the first token ``simple": 

error$\rightarrow$lot$\rightarrow$a$\rightarrow$it$\rightarrow$like$\rightarrow$still$\rightarrow$i$\rightarrow$,$\rightarrow$was $\rightarrow$plot$\rightarrow$ the$\rightarrow$as$\rightarrow$simple

In a hierarchical model, the second clause is compacted into one component,
and the error propagation is  thus given by:

error$\rightarrow$ second-clause $\rightarrow$ 
first-clause
$\rightarrow$ was$\rightarrow$plot$\rightarrow$the$\rightarrow$as$\rightarrow$simple.

Propagation with clause segmentation consists of only 8 operations.
Such a procedure thus tends to attenuate the gradient vanishing problem, 
potentially yielding better performance. 

\end{enumerate}

\begin{figure}
\centering
\includegraphics[width=2.8in]{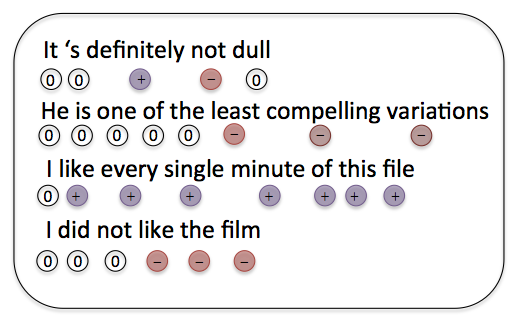}
\caption{Sentiment prediction using a one-directional (left to
right) LSTM. Decisions at each time step are made by feeding embeddings
calculated from the LSTM into a softmax classifier.}
\label{prediction}
\end{figure}

\subsection{Binary Sentiment Classification (Pang)}
\paragraph{Task Description:} The sentiment dataset of \newcite{pang2002thumbs} consists of sentences with a sentiment label for each sentence. 
We divide the original dataset into training(8101)/dev(500)/testing(2000).  
No pre-training procedure as described in \newcite{socher2011semi} is employed.
Word embeddings are initialized using skip-grams
and kept fixed in the learning procedure. 
We trained skip-gram embeddings
on the Wikipedia+Gigaword dataset
 using the word2vec package\footnote{https://code.google.com/p/word2vec/}. 
Sentence level embeddings are fed into a sigmoid classifier.
Performances for 50 dimensional vectors are given in the table below:
 \begin{table}[!ht]
\center
\small
\begin{tabular}{l|l|l}
&Standard&LSTM\\\hline
Tree&0.745&0.774\\\hline\hline
Sequence&0.733 (-0.012)&0.783 (+0.008) \\
P-value&0.060&0.136\\\hline\hline
Bi-Sequence&0.754 (+0.09) &0.790 (+0.016)\\
P-value&0.058&0.024*\\\hline\hline
\end{tabular}
\caption{Test set accuracies on the Pang's sentiment dataset using Standard model settings.}
\end{table}

\paragraph{Discussion}
Why don't parse trees help on this task?  One possible explanation is the
distance of the supervision signal from the local compositional structure.
The Pang et al. dataset has an average sentence length of 22.5 words,
which means it takes multiple steps before sentiment related evidence comes up to the surface.
It is therefore unclear whether local compositional operators (such as negation) can be learned;
there is only a  small amount of training data (around 8,000 examples) and the sentiment supervision only at the level of the sentence
may not be easy to propagate down to deeply buried local phrases.

\subsection{Question-Answer Matching}
\paragraph{Task Description:} In the question-answering dataset 
QANTA\footnote{\url{http://cs.umd.edu/~miyyer/qblearn/}.
Because the publicly released dataset is smaller than the version used in \cite{iyyer2014neural} due to privacy issues,
our numbers are not comparable to those in \cite{iyyer2014neural}.}, each answer is a token or short phrase.
The task
is different from standard generation focused QA task but 
formalized as
a multi-class classification task that matches a source question with a candidates  phrase
from a predefined pool of candidate phrases
We give an illustrative example here:

{\bf Question}: {\it He left unfinished a novel whose title character forges his father's signature to get out of school and avoids the draft by feigning desire to join.  Name this German author of The Magic Mountain and Death in Venice. }

{\bf Answer}: {\it Thomas Mann} from the pool of phrases. Other candidates might include George Washington, Charlie Chaplin, etc.

The model of \newcite{iyyer2014neural} minimizes the distances between answer embeddings and node embeddings along the parse tree of the question. 
Concretely, let $c$ denote the correct answer to question $S$, with embedding $\vec{c}$, and $z$ denoting any random wrong answer. 
The objective function sums over the dot product between representation for every node $\eta$ along the question parse trees and the answer representations:
\begin{equation}
L=\sum_{\eta\in \text{[parse tree]}}\sum_z  max(0, 1-\vec{c}\cdot e_{\eta}+\vec{z}\cdot e_{\eta})
\end{equation}
where $e_{\eta}$ denotes the embedding for parse tree node calculated from the recursive neural model. 
Here the parse trees are dependency parses 
following \cite{iyyer2014neural}.

By adjusting 
the framework 
to recurrent models,
we  minimize the distance between the answer embedding and the embeddings calculated from each timestep $t$ of the sequence:
\begin{equation}
L=\sum_{t\in [1,N_s]}\sum_z  max(0, 1-\vec{c}\cdot e_t+\vec{z}\cdot e_t)
\end{equation}
At test time, the model chooses the answer (from the set of candidates) 
that gives the lowest loss score. 
As can be seen from results presented in Table 5,
the difference is only significant for the LSTM setting between the
tree model and the sequence model; no significant difference is observed for other settings.

 \begin{table}[!ht]
\center
\small
\begin{tabular}{l|l|l}
&Standard&LSTM\\\hline
Tree&0.523&0.558\\\hline\hline
Sequence&0.525 (+0.002)&0.546 (-0.012) \\
P-value&0.490&0.046*\\\hline\hline
Bi-Sequence&0.530 (+0.007) &0.564 (+0.006)\\
P-value&0.075&0.120\\\hline\hline
\end{tabular}
\caption{Test set accuracies for UMD-QA dataset.}
\end{table}

\paragraph{Discussion}
The UMD-QA task represents a group of situations where because we have insufficient
supervision about matching (it's hard to know which node in the parse
tree or which timestep provides the most direct evidence for the answer),
decisions have to be made by looking at and iterating over all
subunits (all nodes in parse trees or timesteps). Similar ideas
can be found in pooling structures (e.g. \newcite{socher2011dynamic}).

The results above illustrate that 
for tasks where 
 we try to align the target with different source components (i.e., parse tree nodes for tree models and different time steps for sequence models), 
components from sequence models are able to embed important information,
despite the fact that sequence model components are just sentence fragments and hence usually not linguistically meaningful components in the way that parse tree constituents are.

\subsection{Semantic Relationship Classification}
\paragraph{Task Description:} SemEval-2010 Task 8 \cite{hendrickx2009semeval} is to find semantic relationships between
pairs of nominals, e.g., in ``My [apartment]$_{\text{e1}}$ has a pretty large [kitchen]$_{\text{e2}}$" 
classifying the relation between [apartment] and [kitchen] as {\it component-whole}. 
The dataset contains 9 ordered relationships, so the task is formalized as a 19-class classification problem,
with directed  relations treated as separate labels; see \newcite{hendrickx2009semeval,socher2012semantic} for details.

For the recursive implementations,
we follow the neural framework defined in \newcite{socher2012semantic}.    
The path in the parse tree between the two nominals is retrieved, and
the embedding is calculated based on recursive models and fed to a softmax classifier\footnote{\cite{socher2012semantic} achieve state-of-art performance by
combining a sophisticated model,  MV-RNN,
in which each word is presented with both a matrix and a vector with
human-feature engineering. Again, because
MV-RNN is difficult to adapt to a recurrent version, we do not
employ this state-of-the-art model, adhering only to the general
versions of recursive models described in Section 2, since our main
goal is to compare equivalent recursive and recurrent models rather than implement the state of the art.}.
Retrieved paths are transformed for the recurrent models as shown in Figure~\ref{relation}.
\begin{figure}[!ht]
\centering
\includegraphics[width=3in]{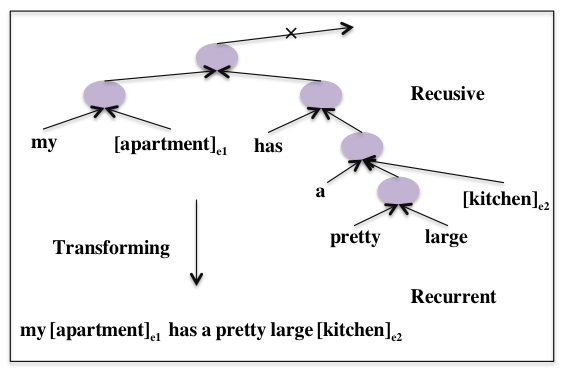}
\caption{Illustration of Models for Semantic Relationship Classification. }
\label{relation}
\end{figure}

\paragraph{Discussion}
Unlike for earlier tasks, 
here recursive models yield much better performance than the corresponding recurrent versions 
for all versions
(e.g., standard tree vs. standard sequence, $p=0.004$).  
These results suggest that it is the need to integrate structures
far apart in the sentence that characterizes the tasks where
recursive models surpass recurrent models. In parse-based  models,
the two target words are drawn together much earlier in the decision
process than in recurrent models, which must remember one target until
the other one appears. 

\begin{table}[!ht]
\center
\small
\begin{tabular}{l|l|l}
&Standard&LSTM\\\hline
Tree&0.748&0.767\\\hline\hline
Sequence&0.712 (-0.036)&0.740 (-0.027) \\
P-value&0.004*&0.020*\\\hline\hline
Bi-Sequence&0.730 (-0.018) &0.752 (-0.014)\\
P-value&0.017*&0.041*\\\hline\hline
\end{tabular}
\caption{Test set accuracies on the SemEval-2010 Semantic Relationship Classification task.}
\end{table}

\subsection{Discourse Parsing}
\begin{figure}[!ht]
\centering
\includegraphics[width=2.1in]{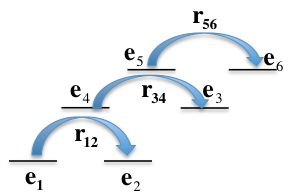}
\caption{{An illustration of discourse parsing. $[e_1, e_2, ...]$ denote EDUs (elementary discourse units), each consisting of a sequence of tokens. $[r_{12}, r_{34}, r_{56}]$ denote relationships to be classified. A binary classification model is first used to decide whether two EDUs should be merged and a multi-class classifier is then used to decide the relation type.}}
\label{relation}
\end{figure}

\paragraph{Task Description:} Our final  task, discourse parsing based on the RST-DT corpus \cite{carlson2003building},
is to build a discourse
tree for a document, based on assigning Rhetorical Structure Theory (RST) relations between
elementary discourse units (EDUs).
Because discourse relations express the coherence structure of discourse,
they presumably express different aspects of compositional meaning than sentiment or 
nominal relations.  
See \newcite{hernault2010hilda} for more details on discourse parsing and the RST-DT corpus.

Representations for adjacent EDUs are fed into binary classification (whether two EDUs are related) 
and multi-class relation classification models, as defined in \newcite{li2014recursive}.
Related EDUs are then merged into a new EDU, the representation of which is obtained through an operation of neural composition based on the previous two related EDUs.
This step is repeated until all units are merged. 

Discourse parsing takes EDUs as the basic units to operate on; EDUs are short clauses, not full sentences, with an average length of 7.2 words. Recursive and recurrent models are applied on EDUs to create embeddings to be used as inputs for discourse parsing. 
We use this task for two reasons:
(1) to illustrate whether syntactic parse trees are useful 
for acquiring  representations for short clauses. 
(2) to measure the extent to which parsing improves discourse tasks that need to combine the meanings of larger text units.

Models are traditionally evaluated in terms of three metrics, i.e., spans\footnote{on 
blank tree structures.}, nuclearity\footnote{on tree structures
with nuclearity indication.},  and identifying the rhetorical relation between two clauses. 
Due to space limits, we only focus the last one, rhetorical relation identification,
because
(1) relation labels are treated as correct only if spans and nuclearity
are correctly labeled
(2) relation identification between clauses offer more insights about model's abilities 
to represent sentence semantics.
In order to perform 
a plain comparison, no additional human-developed features are added.


\begin{table}[!ht]
\center
\small
\begin{tabular}{l|l|l}
&Standard&LSTM\\\hline
Tree&0.568&0.564\\\hline\hline
Sequence&0.572 (+0.004)&0.563 (-0.002) \\
P-value&0.160&0.422\\\hline\hline
Bi-Sequence&0.578 (+0.01) &0.575 (+0.012)\\
P-value&0.054&0.040*\\\hline\hline
\end{tabular}
\caption{Test set accuracies for relation identification on RST discourse parsing data set.}
\end{table}

\paragraph{Discussion}
We see no large differences  
between equivalent recurrent and recursive models. 
We suggest two possible explanations.
(1) EDUs tend to be short; thus for some clauses, 
parsing might not change the order of operations on words.
Even for those whose orders are changed by parse trees,
the influence of short phrases on the final representation may not be great enough.
(2) Unlike earlier tasks, where text representations are immediately used as inputs into classifiers, the algorithm presented here 
adopts additional  levels of neural composition during the process of EDU merging. 
We suspect that neural layers may act as information filters,
separating the informational chaff from the wheat, which in turn makes the model 
a bit more immune to the initial inputs. 


\section{Discussion}

We compared recursive and recurrent
neural models for representation learning on 5 distinct NLP tasks 
in 4 areas for which recursive neural models are known to 
achieve good performance
\cite{socher2012semantic,socher2013recursive,li2014recursive,iyyer2014neural}.

As with any comparison between models, our results come with some caveats:
First, we explore
the most general or basic forms of recursive/recurrent models
rather than various sophisticated algorithm variants.
This is because fair comparison becomes more and more difficult as 
models get complex (e.g., the number of layers, number of hidden units within each layer, etc.).
Thus most neural models employed in this work are comprised of only
one layer of neural compositions---despite the fact that deep neural models with multiple layers give better results.   
Our conclusions  might thus be 
limited to the algorithms employed in this paper,
and it is unclear whether they can be extended to other variants
or to the latest state-of-the-art. 
Second, in order to compare models ``fairly'',
we force
every model to be trained exactly in the same way: AdaGrad with minibatches, same set of initializations, etc. However, this may not
necessarily be the optimal way to train every model;
different training strategies tailored for specific models may
improve their performances.  In that sense, our attempts to be ``fair" 
in this paper may nevertheless be unfair.

Pace these caveats, our conclusions can be summarized as follows:
\begin{itemize}
\item In tasks like semantic relation extraction, in which
single headwords need to be associated across a long distance,
recursive models shine. 
This suggests that for the many other kinds of 
tasks in which long-distance semantic dependencies play a role
(e.g., translation between languages with significant
reordering like Chinese-English translation),
syntactic structures from recursive models may
offer useful power.
\item 
Tree models tend to help more on  long sequences
than shorter ones
 with sufficient supervision: 
tree models slightly help root level identification on 
the Stanford Sentiment Treebank, but do not help much at the phrase level. 
Adopting bi-directional versions of recurrent models seem to
largely bridge this gap, producing equivalent or sometimes better results. 
\item 
On long sequences where
supervision is not sufficient, 
e.g., in Pang at al.,'s dataset (supervision only exists on top of long sequences),
no significant difference is observed between tree based and sequence based models.
\item
In  cases where tree-based models do well,
a simple approximation to tree-based models seems to improve
recurrent models to equivalent or almost equivalent performance:
(1) break long sentences (on punctuation) into a series of clause-like units,
(2) work on these clauses separately, and
(3) join them together. 
This model sometimes works as well as  tree models for the sentiment task, suggesting that 
one of the reasons tree models help is by breaking down long sentences into
more manageable units.
\item Despite that the fact that components (outputs from different time steps) in recurrent models are not linguistically meaningful, they may do as well as linguistically meaningful phrases (represented by parse tree nodes) in embedding informative evidence, as demonstrated in UMD-QA task. 
Indeed, recent work 
in parallel with ours \cite{bowman2015tree}  has shown
that recurrent models like LSTMs can discover implicit recursive compositional structure.
\end{itemize} 


\section{Acknowledgments}
We would especially 
like to thank Richard Socher 
and Kai-Sheng Tai 
for insightful comments, advice, 
and suggestions.  We would also like to thank
Sam Bowman, Ignacio Cases, Jon Gauthier, Kevin Gu,  Gabor Angeli, Sida Wang, Percy Liang 
and other members of the Stanford NLP group, 
as well as the anonymous reviewers for their helpful advice on various aspects of this work. 
We acknowledge the support of NVIDIA Corporation with the donation of Tesla K40 GPUs
We gratefully acknowledge support from
an Enlight Foundation Graduate Fellowship, a gift from Bloomberg L.P., 
the Defense Advanced Research Projects Agency (DARPA) Deep Exploration and Filtering of Text (DEFT) Program under Air Force Research Laboratory (AFRL) contract no. FA8750-13-2-0040,
and the  NSF via award IIS-1514268.
Any opinions, findings, and conclusions or recommendations expressed in this material are those of the authors and do not necessarily reflect the views of Bloomberg L.P., DARPA, AFRL, NSF, or the US government.


\bibliographystyle{acl}
\bibliography{acl2013}

\end{document}